\documentclass[10pt,twocolumn,letterpaper]{article}

\usepackage[pagenumbers]{PaperForReview} 

\usepackage{graphicx}
\usepackage{amsmath}
\usepackage{amssymb}
\usepackage{booktabs}
\usepackage{soul}

\usepackage{algorithm}
\usepackage{algorithmic}

\usepackage[utf8]{inputenc} 
\usepackage[T1]{fontenc}    
\usepackage{url}            
\usepackage{booktabs}       
\usepackage{amsfonts}       
\usepackage{nicefrac}       
\usepackage{microtype}      
\usepackage{xcolor}         

\usepackage{graphicx}       
\usepackage{amsmath}
\usepackage{amssymb}
\usepackage{multicol}
\usepackage{makecell}
\usepackage{multirow}
\usepackage{enumitem,kantlipsum}
\usepackage{wrapfig,lipsum,booktabs}

\newcommand{\tabincell}[2]{\begin{tabular}{@{}#1@{}}#2\end{tabular}}
\newcommand{\mathbbm}[1]{\text{\usefont{U}{bbm}{m}{n}#1}}

\DeclareMathOperator*{\argmin}{arg\,min}

\usepackage[pagebackref,breaklinks,colorlinks]{hyperref}

\usepackage[capitalize]{cleveref}
\crefname{section}{Sec.}{Secs.}
\Crefname{section}{Section}{Sections}
\Crefname{table}{Table}{Tables}
\crefname{table}{Tab.}{Tabs.}

\def\cvprPaperID{3102} 
\def\confName{CVPR}
\def\confYear{2023}

\begin{document}

\title{Open World DETR: Transformer based Open World Object Detection}
\author{
Na Dong$^{1,2}$\thanks{Work fully done while first author is a visiting PhD student at the National University of Singapore.} \qquad Yongqiang Zhang$^2$ \qquad Mingli Ding$^2$ \qquad Gim Hee Lee$^1$ \\
$^1$Department of Computer Science, National University of Singapore\\
$^2$School of Instrument Science and Engineering, Harbin Institute of Technology\\
{\tt\small \{dongna1994, zhangyongqiang, dingml\}@hit.edu.cn \qquad gimhee.lee@comp.nus.edu.sg}
}

\maketitle


\pdfoutput=1

\begin{abstract}
Open world object detection aims at detecting objects that are absent in the object classes of the training data as unknown objects without explicit supervision. 
Furthermore, the exact classes of the unknown objects must be identified without catastrophic forgetting of the previous known classes when the corresponding annotations of unknown objects are given incrementally.
In this paper, we propose a two-stage training approach named Open World DETR for open world object detection based on Deformable DETR. In the first stage, we pre-train a model on the current annotated data to detect objects from the current known classes, and concurrently train an additional binary classifier to classify predictions into foreground or background classes. 
This helps the model to build an unbiased feature representations that can facilitate the detection of unknown classes in subsequent process.
In the second stage, we fine-tune the class-specific components of the model with a multi-view self-labeling strategy and a consistency constraint. 
Furthermore, we alleviate catastrophic forgetting when the annotations of the unknown classes becomes available incrementally by using knowledge distillation and exemplar replay. Experimental results on PASCAL VOC and MS-COCO show that our proposed method outperforms other state-of-the-art open world object detection methods by a large margin. 

\end{abstract}

\vspace{-4mm}
\section{Introduction}
In the past decade, many impressive general object detection methods have been developed due to the huge success of deep learning~\cite{girshick2013rich,uijlings2013selective,girshick2015fast,Lin2017Feature,redmon2015you,Liu2016SSD,lin2017focal,carion2020end, zhu2020deformable}. Despite the tremendous success, current state-of-the-art deep learning based object detection methods have difficulty in recognizing objects of unknown classes as objects because they are designed under a closed world assumption. In other words, given a training dataset with annotations of a set of known classes, the goal of these object detection methods is to detect the known classes while regarding all other objects beyond the given known classes as background. Consequently, these methods are unable to detect unknown objects at inference time. Humans are better than machines in identifying objects of unknown classes as objects, distinguishing unknown objects from the known classes and rapidly recognizing them when their annotations are given incrementally. 
\begin{figure}[t]
\centering
\includegraphics[width=\linewidth,height=5.45cm]{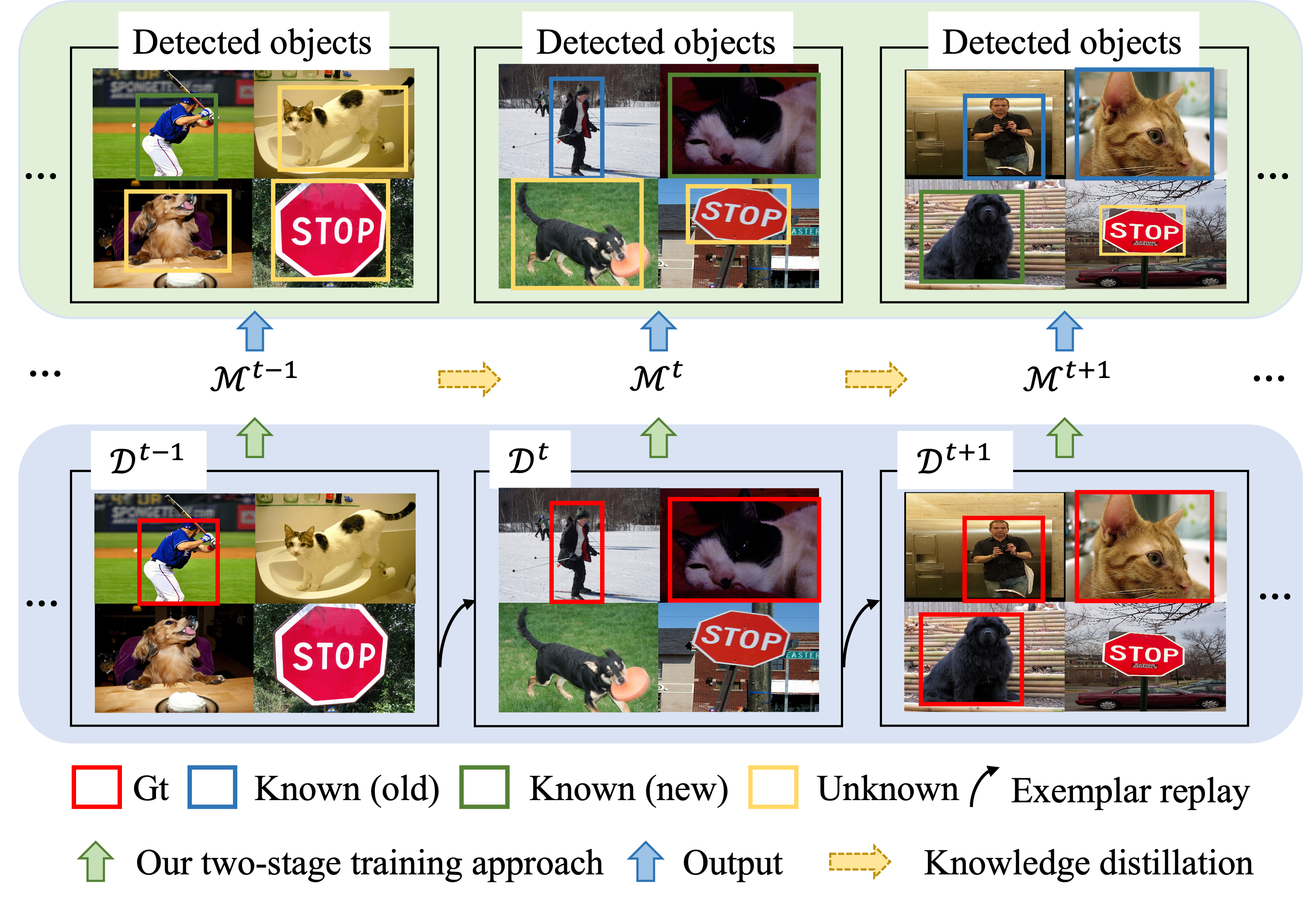}
\caption{Conceptualization of our open world DETR. We propose a two-stage training approach that unifies pre-training and fine-tuning to detect unknown objects as objects without annotations. We train the model $\mathcal{M}^t$ with ground truths of known classes and pseudo ground truths of unknown classes generated by a multi-view self-labeling strategy to detect unknown classes. 
We use knowledge distillation and exemplar replay to mitigate catastrophic forgetting in the incremental learning stage.
} 
\label{model training stage}
\vspace{-5mm}
\end{figure}

Open world object detection aims to circumvent the above mentioned problems from the closed world assumption by training a deep learning-based object detector to detect unknown objects as objects without explicit supervision. Additionally, the unknown objects must be recognized when their annotations are given incrementally, without forgetting the previous known classes.
Specifically, three challenges have to be overcome to achieve open world object detection: 1) In addition to identifying known classes accurately, the model must also learn to generate high-quality proposals for the unknown classes without supervision. 2) A object detection model tends to suppress the potential unannotated unknown objects as background under full supervision from the known classes.
The model must be capable of separating the unknown objects from the background in open world object detection setting. 3) The model trained with annotations of the known classes is incorrectly biased to detecting the unknown objects as one of the known classes. In the open world object detection setting, the model must be able to distinguish the unknown objects from the known classes.

Joseph~\etal~\cite{joseph2021towards} proposes ORE that is built on the two-stage object detection pipeline Faster R-CNN for open world object detection. Specifically, ORE is based on contrastive clustering, an unknown object-aware proposal network, and energy-based unknown identification. 
Although ORE mitigates the challenges in open world object detection, it relies on a held-out validation set for the estimation of the unknown class distribution. In contrast, our method does not need any supervision for the unknown classes and thus agrees better with the definition of open world object detection. Gupta~\etal~\cite{gupta2021ow} introduces another transformer-based open world object detection method OW-DETR that utilizes an attention-driven pseudo-labeling scheme to select object query boxes with high attention scores but not matching any ground truth bounding box from the known classes as pseudo proposals of the unknown classes. Despite the performance improvements over ORE, the reliance of OW-DETR on the attention map results in the localization of only a small part of an unknown object with high attention score while suppressing the rest of the unknown object.

In this paper, we introduce a two-stage training approach based on Deformable DETR~\cite{zhu2020deformable} to do open world object detection. We slightly modify Deformable DETR by adding a class-agnostic binary classification head alongside the existing class-specific classification head and regression head. In the first stage, we pre-train the modified Deformable DETR on the dataset of the current known classes. The additional class-agnostic binary classifier is trained to classify predictions into foreground or background classes, and 
thus alleviating the bias towards the current known classes caused by training on their exact class supervision.
In the second stage, we fine-tune the class-specific projection layer, classification head and the class-agnostic binary classification head on the dataset of current known classes using a multi-view self-labeling strategy and a consistency constraint while keeping the other class-agnostic components of the model frozen.
The multi-view self-labeling strategy fine-tunes these components with pseudo targets of the unknown classes generated by the pre-trained class-agnostic binary classifier and the selective search algorithm~\cite{uijlings2013selective}. The consistency constraint improves the quality of the representations by self-regularization with data augmentation. Furthermore, when the annotations of the unknown classes become available incrementally, we alleviate forgetting of previous known classes by applying knowledge distillation to the outputs of the class-specific components and replaying exemplars.


Our contributions can be summarized as follows: 
\begin{enumerate}[leftmargin=*] 
\item A two-stage training approach which unifies pre-training and fine-tuning is proposed to tackle the challenging and realistic open world setting of object detection.
\item A multi-view self-labeling strategy is introduced to use the class-agnostic binary classifier and the selective search algorithm to generate high-quality pseudo proposals for unknown classes, where swapped prediction is performed. Furthermore, a consistency constraint between the object query features of the two views is used to improve the quality of the feature representations.
\item Extensive experiments conducted on two standard object detection datasets (\ie MS-COCO and PASCAL VOC) demonstrate the significant performance improvement of our approach over existing state-of-the-arts.
\end{enumerate}

\begin{figure*}[!t]
\centering
\includegraphics[width=0.8\linewidth,height=3cm]{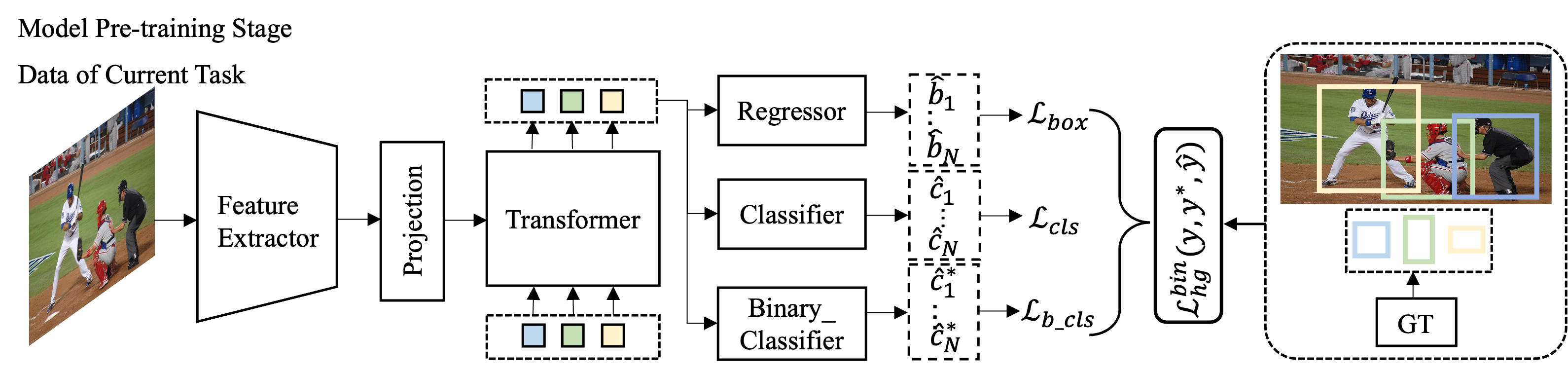}
\caption{Overview of our proposed model pre-training stage. Refer to the text for more details.}
\label{stage 1}
\vspace{-4mm}
\end{figure*}

\vspace{-4mm}
\section{Related Works}
\label{Related Works}

\paragraph{Closed World Object Detection.}
Closed world object detection models work under the assumption that all classes to be detected are available at the training phase.
Existing closed world object detection models generally fall into two categories: 1) two-stage detectors and 2) one-stage detectors. 
Two-stage detectors such as R-CNN~\cite{girshick2013rich} apply a deep neural network to extract features from proposals generated by the selective search algorithm~\cite{uijlings2013selective}. Fast R-CNN~\cite{girshick2015fast}
utilizes a differentiable RoI Pooling to improve the speed and performance. Faster R-CNN~\cite{Ren2015Faster} introduces the Region Proposal Network (RPN) to generate proposals. FPN~\cite{Lin2017Feature} builds a top-down architecture with lateral connections to extract features across multiple layers.
In contrast, one-stage detectors such as YOLO~\cite{redmon2015you} directly perform object classification and bounding box regression on the feature maps. SSD~\cite{Liu2016SSD} uses feature pyramid with different anchor sizes to cover the possible object scales. RetinaNet~\cite{lin2017focal} proposes the focal loss to mitigate the imbalanced positive and negative examples. 
Recently, another category of object detection methods~\cite{carion2020end, zhu2020deformable} beyond the one-stage and two-stage methods have gained popularity. They directly supervise bounding box predictions end-to-end with Hungarian bipartite matching. 
However, these detectors are still constrained by the taxonomy defined by the training dataset. 
Therefore, it is imperative to extend the capability of object detectors to unknown classes undefined by the training data. 

\vspace{-3mm}
\paragraph{Open World Object Detection.}
The open world problem first gained attention in image classification~\cite{scheirer2013towards,bendale2015towards}.
Subsequently, the open world setting has been applied to object detection with the goal of identifying unknown classes on a closed world training dataset. 
Joseph~\etal~\cite{joseph2021towards} proposes ORE, an open world object detector based on the two-stage object detection pipeline Faster R-CNN~\cite{Ren2015Faster}. ORE utilizes an unknown object-aware RPN to generate proposals for the unknown classes. Furthermore, ORE learns an energy-based classifier to distinguish the unknown classes from the known classes. However, ORE relies on a held-out validation set with annotations of unknown classes to estimate the distribution of unknown classes which violates the definition of open world object detection. Gupta~\etal~\cite{gupta2021ow} proposes OW-DETR, an end-to-end transformer-based framework for open world object detection.
An attention-driven pseudo-labeling scheme is used to select pseudo proposals for the unknown classes. 
Furthermore, an objectness branch is used to learn a separation between foreground objects and the background by enabling knowledge transfer from known classes to the unknown classes.  
In contrast, we propose to collaboratively use a class-agnostic binary classifier and the selective search algorithm to generate more accurate and reliable pseudo proposals for the unknown classes. 


\vspace{-2mm}
\section{Problem Definition}
Let $\mathcal{K}^t = \{ 1, 2, \dots, C \}$ denote the set of current known classes at time $t$. Furthermore, let $\mathcal{U}^t = \{C+1, \dots \} $ denote a set of unknown classes which may be encountered during inference at time $t$.  Let $\mathcal{D}^t = \mathit{\{x^t,y^t\}}$ denote a dataset for the current known classes at time t, which contains images $\mathit{x^t}$ and corresponding ground truths $\mathit{y^t}$.
Following the definition of open world object detection setting, at time $t$, an object detector $\mathcal{M}^t$ is trained to: 1) identify objects belonging to the current known classes $\mathcal{K}^t$; 2) detect objects of the unknown classes $\mathcal{U}^t$ as objects; 3) recognize the previously encountered known classes. 
Subsequently, a set of $n$ new classes from the unknown classes $\mathcal{U}^t$ and their training dataset with corresponding annotations $\mathcal{D}^{t+1}=\mathit{\{x^{t+1},y^{t+1}\}}$ are given at time $t+1$. 
The known classes and unknown classes are updated as $\mathcal{K}^{t+1} = \{ 1, 2, \dots, C, C+1, \dots, C+n \}$ and $\mathcal{U}^{t+1} = \{C+n+1, \dots\} $, respectively. Additionally, the object detector $\mathcal{M}^t$ must be updated to $\mathcal{M}^{t+1}$ without retraining from scratch on the whole dataset.

\begin{figure*}[!t]
\centering
\includegraphics[width=\linewidth,height=7.6cm]{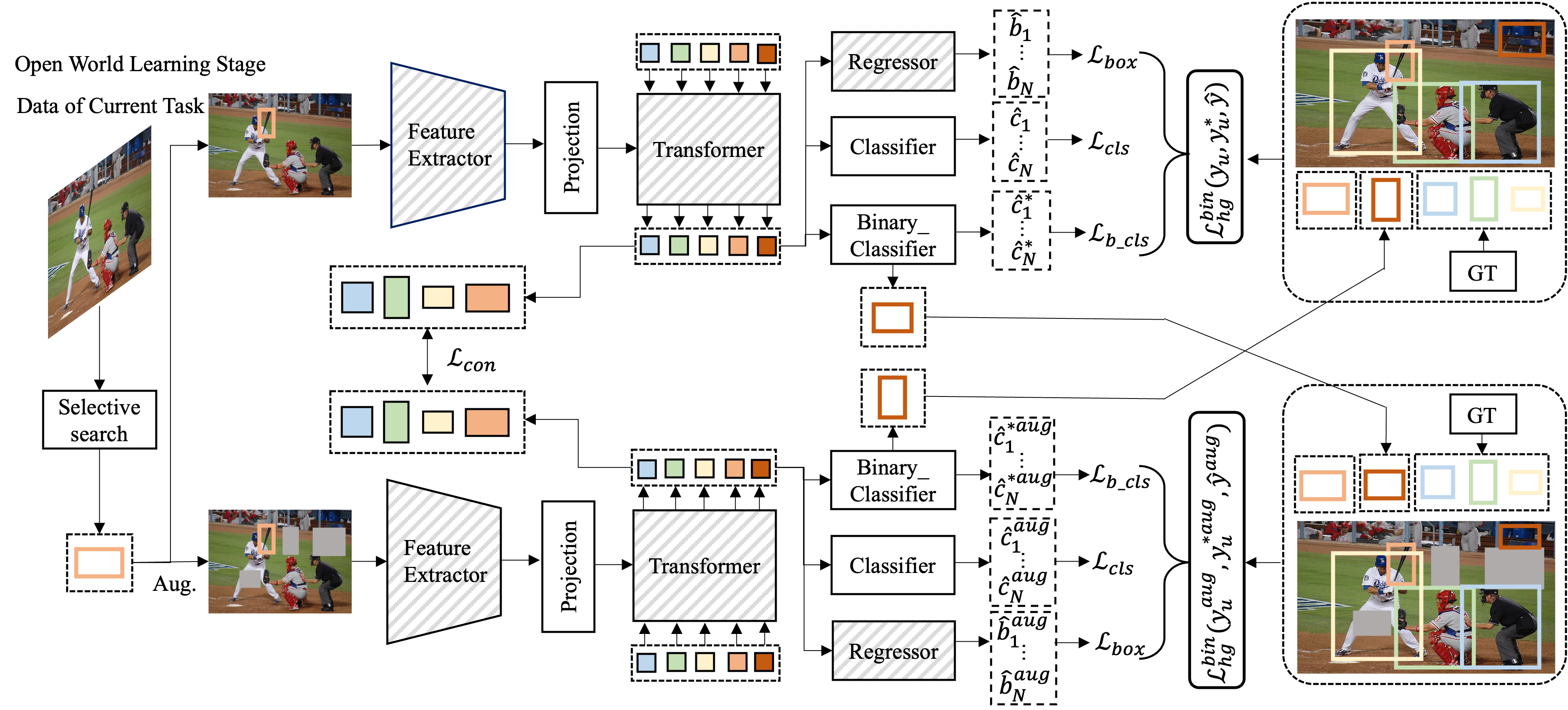}
\caption{Overview of our proposed open world learning stage. Parameters of the modules shaded in gray slash line are frozen during training. Note that there is only one network. Our illustration contains two networks to clearly show the swapped prediction mechanism for our proposed multi-view self-labeling strategy. The two sets of loss terms are instances of the same objective applied to different views.} 
\label{stage 2}
\vspace{-4mm}
\end{figure*}

\vspace{-2mm}
\section{Our Approach}
\label{Proposed Approach}


\subsection{Model Pre-training Stage}

When trained with full supervision of the current known classes, Deformable DETR can only classify its predictions into one of the current known classes or background and would fail to detect the potential unknown objects as objects. 
To circumvent this problem, we propose to add an additional \textbf{class-agnostic binary classification head} in Deformable DETR and train it to classify all predictions into either foreground or background classes.
Since the binary classifier is trained without the need to classify the predictions into the exact object class, the model should be free from bias towards the current known classes. 
Therefore, the model can build unbiased feature representations which facilitates detecting unknown objects as objects in subsequent process.

As shown in Figure~\ref{stage 1}, we train the full model that consists of the additional class-agnostic binary classification head alongside the existing class-specific classification head and regression head on the dataset $\mathcal{D}^t$ with annotations of the current known classes $\mathcal{K}^t$.
Following Deformable DETR, we denote the set of $N$ predictions of the current known classes as $\hat{y} = \{ \hat{y}_i \}^N_{i=1} = \{  (\hat{c}_i, \hat{\mathbf{b}}_i)\}^N_{i=1}$, and the ground truth set of objects as $y = \{ y_i \}^N_{i=1} = \{(c_i, \mathbf{b}_i)\}^N_{i=1}$ padded with $\varnothing$ (no object). 
For each element $i$ of the ground truth set, $c_i$ is the target class label (which may be $\varnothing$ ) and $\mathbf{b}_i \in [0, 1]^4$ is a 4-vector that defines ground truth box center coordinates, and its height and width relative to the image size. 

Since the matching cost of Deformable DETR takes into account both the class predictions and the similarity of the predicted and ground truth boxes, we combine the class-agnostic binary classification head and the regression head to search for a optimal matching. Therefore, for the binary classification head, we further denote the set of $N$ predictions as $\hat{y}^* = \{ \hat{y}^*_i \}^N_{i=1} = \{  (\hat{c}^*_i, \hat{\mathbf{b}}_i)\}^N_{i=1}$.
The ground truths are denoted as $y^* = \{ y^*_i \}^N_{i=1} = \{(c^*_i, \mathbf{b}_i)\}^N_{i=1}$, where $c^*_i \in \{0, 1\}$, $0$ denotes foreground class, \ie the current known classes, and $1$ denotes the background class.
During training, we collectively assign the predictions that match the ground truths of the current known classes with the foreground class as label $0$, and the predictions of the background with the background class as label $1$.

We adopt a pair-wise matching cost $\mathcal{L}_\text{match}(y_i, \hat{y}_{\sigma(i)})$ between ground truth $y_i$ and a prediction $\hat{y}_{\sigma(i)}$ with index $\sigma(i)$ to search for a  bipartite matching with the lowest cost. Concurrently, we also adopt a pair-wise matching cost $\mathcal{L}_\text{match}(y^*_i, \hat{y}^*_{\sigma^*(i)})$ between ground truth $y^*_i$ and a prediction $\hat{y}_{\sigma^*(i)}$ with index $\sigma^*(i)$ to search for a bipartite matching with the lowest cost, \ie
\begin{equation}
\small
\begin{array}{rll}
\begin{aligned}
\displaystyle 
\mathcal{\hat{\sigma}} &= \argmin_{\sigma} \sum^N_{i=1} \mathcal{L}_\text{match}(y_i, \hat{y}_{\sigma(i)}),\\
\mathcal{\hat{\sigma}^*} &= \argmin_{\sigma^*} \sum^N_{i=1} \mathcal{L}_\text{match}(y^*_i, \hat{y}^*_{\sigma^*(i)}).
\end{aligned}
\end{array}
\end{equation}

The matching cost of the existing classification and regression heads is defined as:
\begin{equation}
\small
\begin{array}{rll}
\begin{aligned}
\displaystyle 
\mathcal{L}_\text{match}(y_i, \hat{y}_{\sigma(i)}) &= \mathbbm{1}_{\{c_i \neq \varnothing \}} \mathcal{L}_\text{cls}(c_i, \hat{c}_{\sigma(i)}) \\ & +  \mathbbm{1}_{\{c_i \neq \varnothing \}} \mathcal{L}_\text{box}(\mathbf{b}_i, \hat{\mathbf{b}}_{\sigma(i)}).
\end{aligned}
\label{eq2}
\end{array}
\end{equation}

The matching cost of the binary classification head and the regression head is defined as:
\begin{equation}
\small
\begin{array}{rll}
\begin{aligned}
\displaystyle 
\mathcal{L}_\text{match}(y^*_i, \hat{y}_{\sigma^*(i)}) &= \mathbbm{1}_{\{c^*_i \neq \varnothing \}} \mathcal{L}_\text{binary\_cls}(c^*_i, \hat{c}^*_{\sigma^*(i)})  \\ &+  \mathbbm{1}_{\{c^*_i \neq \varnothing \}} \mathcal{L}_\text{box}(\mathbf{b}_i, \hat{\mathbf{b}}_{\sigma^*(i)}).
\end{aligned}
\label{eq3}
\end{array}
\end{equation}

Given the above definitions, the Hungarian loss \cite{lin2017focal} for all pairs matched in the previous step is defined as:
\begin{equation}
\small
\begin{array}{rll}
\begin{aligned}
\displaystyle 
&\mathcal{L}^\text{bin}_\text{hg} (y, y^*, \hat{y}) = \sum_{i=1}^N [\mathcal{L}_\text{cls}(c_i, \hat{c}_{\hat{\sigma}(i)})\\ & +  \mathbbm{1}_{\{c_i \neq \varnothing \}} \mathcal{L}_\text{box}(\mathbf{b}_i, \hat{\mathbf{b}}_{\hat{\sigma}(i)}) + \lambda_\text{b\_cls} \mathcal{L}_\text{b\_cls}(c^*_i, \hat{c}^*_{\hat{\sigma}^*(i)})],
\end{aligned}
\end{array}\label{eq:hungarianBin}
\end{equation}
where $\hat{\sigma}$ and $\hat{\sigma}^*$ denote the optimal assignments computed in Equations~\ref{eq2} and~\ref{eq3}, respectively. $\mathcal{L}_\text{cls}$ and $\mathcal{L}_\text{b\_cls}$ are the sigmoid focal loss \cite{lin2017focal}. $\mathcal{L}_\text{box}$ is a linear combination of $\ell_1$ loss and generalized IoU loss \cite{rezatofighi2019generalized} with the same weight hyperparameters as Deformable DETR. $\lambda_\text{b\_cls}$ is the hyperparameter to balance the loss terms.


\subsection{Open World Learning Stage}

Based on the results of our empirical experiments, we identify that the projection layer and classification head are class-specific and the CNN backbone, transformer and regression head of Deformable DETR are class-agnostic.
We thus propose to fine-tune the class-specific components while keeping the class-agnostic components frozen after initializing the parameters of the model from the pre-trained model. Fine-tuning the class-specific components while keeping the class-agnostic components frozen benefits the model in detecting novel classes of objects without catastrophically forgetting the previous ones.

\vspace{-3mm}
\paragraph{Multi-view self-labeling.} 
To address the missing annotations of the unknown classes, we propose a multi-view self-labeling strategy to generate pseudo ground truths of the unknown classes for training the model to detect unknown objects as objects. Figure~\ref{stage 2} shows an illustration of our multi-view self-labeling strategy. We first adopt a common data augmentation technique of applying random cropping and resizing to image $\mathcal{I}$ to get an augmented view denoted by $\mathcal{I}^\text{aug}$. 
We denote the sets of $N$ predictions of $\mathcal{I}$ and  $\mathcal{I}^\text{aug}$ as $\hat{y} = \{ \hat{y}_{i} \}^N_{i=1} = \{  (\hat{c}_i, \hat{\mathbf{b}}_{i})\}^N_{i=1}$, and $\hat{y}^\text{aug} = \{ \hat{y}_i^\text{aug} \}^N_{i=1} = \{  (\hat{c}_i^\text{aug}, \hat{\mathbf{b}}_i^\text{aug})\}^N_{i=1}$, respectively.

We then rely on the class-agnostic binary classifier to select pseudo proposals for the unknown classes from the predictions.
We apply confidence-based filtering to each prediction with a threshold $\delta$ to filter low-confidence predicted proposals.
To avoid duplicated proposal predictions on the same object, 
we apply non-maximum suppression (NMS) before the use of confidence-based filtering. 
However, significant overlaps of the pseudo proposals with the ground truth objects of the current known classes can degenerate the performance of the current known classes. 
We alleviate the issue by choosing only the top scoring proposals for the unknown classes that are not overlapping with the ground truth objects of the current known classes.

A swapped prediction mechanism is performed based on the two views to encourage the model in making consistent predictions for different views of the same image. Specifically, $\mathcal{I}$ is used to generate the pseudo ground truths $y'$ for $\mathcal{I}^\text{aug}$, and $\mathcal{I}^\text{aug}$ is used to generate the pseudo ground truths ${y'}^\text{aug}$ for $\mathcal{I}$. 
Let $c'$ denotes the pseudo label for the selected pseudo proposal $\mathbf{b}'$ of an unknown object.
Furthermore, let $n$ denote the total number of known classes, we thus set $c’$ as $n+1$ with the ground truth labels of known classes taking $1, \cdots, n$.
Therefore, the pseudo ground truth set of image $\mathcal{I}$ is ${y'}^\text{aug} = \{ {y'}_i^\text{aug} \}^N_{i=1}= \{  ({c'}_i^\text{aug}, {\mathbf{b}'}_i^\text{aug})\}^N_{i=1}$ (${c'}^\text{aug}=c'$) and the pseudo ground truth set of the image $\mathcal{I}^\text{aug}$ is $ y' = \{ y'_i \}^N_{i=1} = \{  (c'_i, {\mathbf{b}'}_i)\}^N_{i=1}$. Finally, we use the pseudo ground truths of the unknown classes combined with the ground truths of current known classes as supervision to train the model, i.e. for image $\mathcal{I}$, the unified training target is $y_u = [y, {y'}^\text{aug}]$, and for image $\mathcal{I}^\text{aug}$, the unified training target is $y_u^{aug} = [y, y']$. 

Concurrently, we keep updating the binary classifier.
For image $\mathcal{I}$, we denote the unified training target of the binary classifier as $y^*_u$. For image $\mathcal{I}^\text{aug}$, we denote the unified training target of the binary classifier as $y_u^{*aug}$. Similarly, during training, we assign the predictions that match the ground truths of the current known classes and the pseudo ground truths of unknown classes with the foreground class as label $0$, and the predictions of the background with the background class as label $1$.

The Hungarian losses for training the same image in two different views are defined as $\mathcal{L}^\text{bin}_\text{hg} (y_u, y_u^*, \hat{y})$ and $\mathcal{L}^\text{bin}_\text{hg}(y_u^\text{aug}, {y_u^*}^\text{aug}, \hat{y}^\text{aug})$, respectively.


\vspace{-3mm}
\paragraph{Supplementary pseudo proposals.}
We propose to generate additional proposals using selective search algorithm~\cite{uijlings2013selective} as a supplement for the pseudo proposals of unknown classes. Specifically, we select the proposals generated by selective search that are not overlapping with the ground truth objects of the current known classes and also not overlapping with the pseudo ground truths generated by the binary classifier as additional pseudo proposals for the unknown classes.

\vspace{-3mm}
\paragraph{Consistency constraint.}
To enhance the quality of the feature representations, we propose to apply a consistency constraint between the object query features of $\mathcal{I}$ and $\mathcal{I}^\text{aug}$.
Note that the pseudo ground truths generated by the binary classifier for $\mathcal{I}$ do not correspond to the pseudo ground truths generated by the binary classifier for $\mathcal{I}^\text{aug}$. Therefore, we only apply consistency constraint on the object query features matched the ground truths of the known classes and the pseudo ground truths generated by selective search.
We first adopt a pair-wise matching loss to find the optimal assignments $\hat{\sigma}$ and $\hat{\sigma}^\text{aug}$ of $\mathcal{I}$ and $\mathcal{I}^\text{aug}$, respectively.
The consistency loss between the matched object query features then given as follows:
\begin{equation}
\small
\begin{array}{rll}
\begin{aligned}
\displaystyle 
\mathcal{L}_\text{con} (\hat{q}, \hat{q}^\text{aug}) = \sum_{i=1}^M \ell_1(\hat{q}_{\hat{\sigma}(i)}, \hat{q}^\text{aug}_{\hat{\sigma}^\text{aug}(i)}),
\end{aligned}
\end{array}
\end{equation}
where $\hat{q}$ and $\hat{q}^\text{aug}$ denote the object-dependent query features of $\mathcal{I}$ and $\mathcal{I}^\text{aug}$ output by the transformer decoder of Deformable DETR.


The overall loss $\mathcal{L}_\text{total}$ of the open world learning stage is given by:
\begin{equation}
\begin{array}{rll}
\begin{aligned}
\displaystyle 
\mathcal{L}_\text{total} &= \mathcal{L}^\text{bin}_\text{hg} (y_u, y_u^*, \hat{y}) + \mathcal{L}^\text{bin}_\text{hg}(y_u^\text{aug}, {y_u^*}^\text{aug}, \hat{y}^\text{aug}) \\ &+ \lambda_\text{con} \mathcal{L}_\text{con} (\hat{q}, \hat{q}^\text{aug}),
\end{aligned}
\end{array}
\end{equation}
where $\mathcal{L}^\text{bin}_\text{hg} (y_u, y_u^*, \hat{y})$ and $\mathcal{L}^\text{bin}_\text{hg}(y_u^\text{aug}, {y_u^*}^\text{aug}, \hat{y}^\text{aug})$
are the Hungarian losses for all pairs matched of the existing classification head, regression head and the binary classification head for $\mathcal{I}$ and $\mathcal{I}^\text{aug}$ as defined in Equation~\ref{eq:hungarianBin}.
$\lambda_\text{con}$ is the hyperparameter to balance the loss terms.

\vspace{-3mm}
\paragraph{Alleviating catastrophic forgetting.}
In addition to detecting unknown objects as objects, the model is required to overcome forgetting of the previous known classes when trained only on the dataset with annotations of current known classes. 
There are many methods proposed for tackling this issue, including exemplar replay~\cite{castro2018end,chaudhry2018efficient,lopez2017gradient,rebuffi2017icarl}, knowledge distillation~\cite{shmelkov2017incremental,kj2021incremental,dong2021bridging}, \etc 
We propose to use both knowledge distillation and exemplar replay to mitigate the forgetting of previous known classes. 
Specifically, knowledge distillation is applied to features and classification outputs when training on the dataset of current known classes in each incremental step. 
Additionally, a balanced set of exemplars of the previous known classes and the current known classes is stored, and the model is fine-tuned on these data after the incremental step. More details can be found in our supplementary materials. For brevity, we omit the drawing of the knowledge distillation pipeline in Figures~\ref{stage 1} and~\ref{stage 2}. 

\vspace{-2mm}
\section{Experiments}
\subsection{Experimental Setup}

\begin{table}[!t]
\small
\centering
\resizebox{\linewidth}{!}{
\begin{tabular}{c|cccc}
\hline
       Task IDs        & Task 1      & Task 2                                                                           & Task 3                                                 & Task 4                                                                           \\ \hline
\begin{tabular}[c]{@{}c@{}}Semantic \\ split\end{tabular} & \begin{tabular}[c]{@{}c@{}} VOC \\ Classes \end{tabular} & \begin{tabular}[c]{@{}c@{}}Outdoor,\\ Accessories,\\ Appliance, \\Truck\end{tabular} & \begin{tabular}[c]{@{}c@{}}Sports,\\ Food\end{tabular} & \begin{tabular}[c]{@{}c@{}}Electronic, \\ Indoor,\\ Kitchen, \\Furniture\end{tabular} \\ \hline
Train set      & 16,551       & 45,520                                                                            & 39,402                                                  & 40,260                                                                            \\ 
Test set       & 4,952        & 1,914                                                                             & 1,642                                                   & 1,738                                                                             \\ \hline
\end{tabular}}
\caption{\small Task composition in the proposed Open World DETR evaluation protocol.} 
\vspace{-4mm}
\label{task composition}
\end{table}

\paragraph{Datasets.}
We follow the data setups of prior works~\cite{joseph2021towards,gupta2021ow} on open world object detection. Specifically, we conduct the experimental evaluations on two widely used object detection benchmarks MS-COCO~\cite{lin2014microsoft} and PASCAL VOC~\cite{everingham2010pascal}. Classes are grouped into a set of tasks $\mathcal{T} = \{\mathcal{T}_1, \dots, \mathcal{T}_t, \dots \}$. There are no overlapping classes between these tasks. All the classes in each corresponding task $\mathcal{T}_t$ is not given until time $t$ is reached. While learning a task $\mathcal{T}_t$ at time $t$, all classes of tasks $\{\mathcal{T}_{\tau}: \tau \leq t\}$ are taken as known classes, and all the classes of tasks $\{\mathcal{T}_{\tau}: \tau > t\}$ are taken as unknown classes. MS-COCO contains objects from 80 different classes, including 20 classes that overlap with PASCAL VOC. As in~\cite{joseph2021towards,gupta2021ow}, these classes are split into 4 tasks as shown in Table~\ref{task composition}. The training set for the first task is from PASCAL VOC and the training sets for the next three tasks are selected from MS-COCO. All images from the \textit{test} set of PASCAL VOC and the selected images from the \textit{val} set of MS-COCO constitute the evaluation set, where all previous known classes and current known classes are labeled, and all classes of future tasks will be labeled "unknown". 

\vspace{-3mm}
\paragraph{Inference.}
Let $n$ be the number of both previous and current known classes at time $t$, the label "unknown" of all the unknown classes are set as $n+1$ at inference time. Since we set the label of the pseudo ground truths as $n+1$ at training time, setting the label of all the unknown classes as $n+1$ at inference time guarantees that the unknown classes can be detected as "unknown" during inference.

\vspace{-3mm}
\paragraph{Metrics.}
Following~\cite{joseph2021towards,gupta2021ow},
we report the PASCAL VOC style (mAP [0.5]) accuracy for known classes. For the unknown classes, the Unknown Recall (U-Recall) is utilized as the main evaluation metric that measures the ability of the model to retrieve the unknown instances. 
Additionally, Wilderness Impact (WI) and Absolute Open-Set Error (A-OSE) are also reported. The Wilderness Impact (WI) metric measures the confusion of the model in predicting an unknown instance as one of the known classes. On the other hand, the Absolute Open-Set Error (A-OSE) metric measures the total number of unknown instances detected as the known classes.

\begin{table*}[!t]
\Large
\centering
\resizebox{\linewidth}{!}{
\begin{tabular}{c|cc|cccc|cccc|ccc}
\hline
Task IDs                                                          & \multicolumn{2}{c|}{Task 1}                                                                              & \multicolumn{4}{c|}{Task 2}                                                                                                                                                                                            & \multicolumn{4}{c|}{Task 3}                                                                                                                                                                                            & \multicolumn{3}{c}{Task 4}                                                                                                                                                                                            \\ \hline
                                                                  & \multicolumn{1}{c}{U-Recall}                  & mAP($\uparrow$)                            & \multicolumn{1}{c}{U-Recall}                  & \multicolumn{3}{c|}{mAP($\uparrow$)}                                                                                                                     & \multicolumn{1}{c}{U-Recall}                  & \multicolumn{3}{c|}{mAP($\uparrow$)}                                                                                                                                 & \multicolumn{3}{c}{mAP($\uparrow$)}                                                                                                                     \\ 
                                                                  & \multicolumn{1}{c}{($\uparrow$)} & \begin{tabular}[c]{@{}c@{}}Current\\ known\end{tabular} & \multicolumn{1}{c}{($\uparrow$)} & \multicolumn{1}{c}{\begin{tabular}[c]{@{}c@{}}Previously\\ known\end{tabular}} & \multicolumn{1}{c}{\begin{tabular}[c]{@{}c@{}}Current\\ known\end{tabular}} & Both & \multicolumn{1}{c}{($\uparrow$)} & \multicolumn{1}{c}{\begin{tabular}[c]{@{}c@{}}Previously\\ known\end{tabular}} & \multicolumn{1}{c}{\begin{tabular}[c]{@{}c@{}}Current\\ known\end{tabular}} & Both  & \multicolumn{1}{c}{\begin{tabular}[c]{@{}c@{}}Previously\\ known\end{tabular}} & \multicolumn{1}{c}{\begin{tabular}[c]{@{}c@{}}Current\\ known\end{tabular}} & Both \\ \hline
Faster-RCNN~\cite{Ren2015Faster}$^{\dag}$                                                       &  \multicolumn{1}{c|}{-}                          & 56.4                                                        & \multicolumn{1}{c|}{-}                          &  \multicolumn{1}{c|}{3.7}                                                           & \multicolumn{1}{c|}{26.7}                                                        & 15.2     & \multicolumn{1}{c|}{-}                          & \multicolumn{1}{c|}{2.5}                                                           & \multicolumn{1}{c|}{15.2}                                                        & 6.7     & \multicolumn{1}{c|}{0.8}                                                                                                                         & \multicolumn{1}{c|}{14.5}                                                        &4.2      \\ 
\begin{tabular}[c]{@{}c@{}}Faster-RCNN+Finetuning$^{\dag}$\end{tabular} & \multicolumn{1}{c|}{-}                          & -                                                        & \multicolumn{1}{c|}{-}                          & \multicolumn{1}{c|}{51.0}                                                           & \multicolumn{1}{c|}{25.0}                                                        &38.0      & \multicolumn{1}{c|}{-}                          & \multicolumn{1}{c|}{38.2}                                                           & \multicolumn{1}{c|}{13.6}                                                        & 30.0     & \multicolumn{1}{c|}{29.7}                                                                                                                          & \multicolumn{1}{c|}{13.0}                                                        &25.6      \\ 
DDETR~\cite{zhu2020deformable}                                                            & \multicolumn{1}{c|}{-}                          &60.4                                                         & \multicolumn{1}{c|}{-}                          & \multicolumn{1}{c|}{3.9}                                                           & \multicolumn{1}{c|}{38.0}                                                        &21.0      & \multicolumn{1}{c|}{-}                          & \multicolumn{1}{c|}{1.2}                                                           & \multicolumn{1}{c|}{26.5}                                                        & 9.6    & \multicolumn{1}{c|}{1.8}                                                                                                                          & \multicolumn{1}{c|}{22.5}                                                        & 7.0     \\ 
\begin{tabular}[c]{@{}c@{}}DDETR+Finetuning\end{tabular}       & \multicolumn{1}{c|}{-}                          &-                                                         & \multicolumn{1}{c|}{-}                          & \multicolumn{1}{c|}{54.1}                                                           & \multicolumn{1}{c|}{36.1}                                                        & 45.1     & \multicolumn{1}{c|}{-}                          & \multicolumn{1}{c|}{40.1}                                                           & \multicolumn{1}{c|}{24.0}                                                        & 34.7     & \multicolumn{1}{c|}{33.1}                                                                                                                         & \multicolumn{1}{c|}{19.9}                                                        & 29.8     \\ \hline
ORE-EBUI~\cite{joseph2021towards}$^{\dag}$                                                         & \multicolumn{1}{c|}{4.9}                          & 56.0                                                        & \multicolumn{1}{c|}{2.9}                          & \multicolumn{1}{c|}{52.7}                                                           & \multicolumn{1}{c|}{26.0}                                                        & 39.4     & \multicolumn{1}{c|}{3.9}                          & \multicolumn{1}{c|}{38.2}                                                           & \multicolumn{1}{c|}{12.7}                                                        &29.7      & \multicolumn{1}{c|}{29.6}                                                                                                                      & \multicolumn{1}{c|}{12.4}                                                        &25.3      \\ 
OW-DETR~\cite{gupta2021ow}                                                          & \multicolumn{1}{c|}{7.5}                          &59.2                                                        & \multicolumn{1}{c|}{6.2}                          & \multicolumn{1}{c|}{\textbf{53.6}}                                                           & \multicolumn{1}{c|}{33.5}                                                        & 42.9     & \multicolumn{1}{c|}{5.7}                          & \multicolumn{1}{c|}{38.3}                                                           & \multicolumn{1}{c|}{15.8}                                                        & 30.8    & \multicolumn{1}{c|}{31.4}                                                                                                                          & \multicolumn{1}{c|}{17.1}                                                        &27.8     \\ 
Ours                                                              & \multicolumn{1}{c|}{\textbf{21.0}}                          &\textbf{59.9}                                                         & \multicolumn{1}{c|}{\textbf{15.7}}                          & \multicolumn{1}{c|}{51.8}                                                           & \multicolumn{1}{c|}{\textbf{36.4}}                                                        & \textbf{44.1}     & \multicolumn{1}{c|}{\textbf{17.4}}                          & \multicolumn{1}{c|}{\textbf{38.9}}                                                           & \multicolumn{1}{c|}{\textbf{24.7}}                                                        & \textbf{34.2}     & \multicolumn{1}{c|}{\textbf{32.0}}                                                                                                                      & \multicolumn{1}{c|}{\textbf{19.7}}                                                        & \textbf{29.0}     \\ \hline
\end{tabular}}
\caption{\small Comparison on MS-COCO. Results of methods marked $\dag$ are from OW-DETR~\cite{gupta2021ow}. Task 1: No fine-tuning since no previous known classes. 
No U-Recall for Faster R-CNN and Deformable DETR (DDETR) since they are not open world object detector. Task 4: No U-Recall since all 80 classes are known.}
\label{results 1}
\end{table*}

\begin{table*}[!htbp]
\huge
\centering
\resizebox{\linewidth}{!}{
\begin{tabular}{c|ccc|ccccc|ccccc|ccc}
\hline
Task IDs                                                          & \multicolumn{3}{c|}{Task 1}                                                                                                   & \multicolumn{5}{c|}{Task 2}                                                                                                                                                                                                                 & \multicolumn{5}{c|}{Task 3}                                                                                                                                                                                                                 & \multicolumn{3}{c}{Task 4}                                                                                                                                           \\ \hline
                                                                  & \multicolumn{1}{c}{WI}          & \multicolumn{1}{c}{A-OSE}       & mAP($\uparrow$)                                            & \multicolumn{1}{c}{WI}          & \multicolumn{1}{c}{A-OSE}       & \multicolumn{3}{c|}{mAP($\uparrow$)}                                                                                                                                     & \multicolumn{1}{c}{WI}          & \multicolumn{1}{c}{A-OSE}       & \multicolumn{3}{c|}{mAP($\uparrow$)}                                                                                                                                     & \multicolumn{3}{c}{mAP($\uparrow$)}                                                                                                                                     \\ 
                                                                  & \multicolumn{1}{c}{($\downarrow$)} & \multicolumn{1}{c}{($\downarrow$)} & \begin{tabular}[c]{@{}c@{}}Current\\ known\end{tabular} & \multicolumn{1}{c}{($\downarrow$)} & \multicolumn{1}{c}{($\downarrow$)} & \multicolumn{1}{c}{\begin{tabular}[c]{@{}c@{}}Previously\\ known\end{tabular}} & \multicolumn{1}{c}{\begin{tabular}[c]{@{}c@{}}Current\\ known\end{tabular}} & Both & \multicolumn{1}{c}{($\downarrow$)} & \multicolumn{1}{c}{($\downarrow$)} & \multicolumn{1}{c}{\begin{tabular}[c]{@{}c@{}}Previously\\ known\end{tabular}} & \multicolumn{1}{c}{\begin{tabular}[c]{@{}c@{}}Current\\ known\end{tabular}} & Both & \multicolumn{1}{c}{\begin{tabular}[c]{@{}c@{}}Previously\\ known\end{tabular}} & \multicolumn{1}{c}{\begin{tabular}[c]{@{}c@{}}Current\\ known\end{tabular}} & Both \\ \hline
Faster-RCNN~\cite{Ren2015Faster}$^{\dag}$                                                     & \multicolumn{1}{c|}{0.0699}            & \multicolumn{1}{c|}{13396}            & 56.4                                                        & \multicolumn{1}{c|}{0.0371}            & \multicolumn{1}{c|}{12291}            & \multicolumn{1}{c|}{3.7}                                                           & \multicolumn{1}{c|}{26.7}                                                        & 15.2     & \multicolumn{1}{c|}{0.0213}            & \multicolumn{1}{c|}{9174}            & \multicolumn{1}{c|}{2.5}                                                           & \multicolumn{1}{c|}{15.2}                                                        & 6.7     & \multicolumn{1}{c|}{0.8}                                                           & \multicolumn{1}{c|}{14.5}                                                        & 4.2     \\ 
\begin{tabular}[c]{@{}c@{}}Faster-RCNN +Finetuning$^{\dag}$\end{tabular} & \multicolumn{1}{c|}{-}            & \multicolumn{1}{c|}{-}            &  -                                                       & \multicolumn{1}{c|}{0.0375}            & \multicolumn{1}{c|}{12497}            & \multicolumn{1}{c|}{51.0}                                                           & \multicolumn{1}{c|}{25.0}                                                        &38.0      & \multicolumn{1}{c|}{0.0279}            & \multicolumn{1}{c|}{9622}            & \multicolumn{1}{c|}{38.2}                                                           & \multicolumn{1}{c|}{13.6}                                                        & 30.0     & \multicolumn{1}{c|}{29.7}                                                           & \multicolumn{1}{c|}{13.0}                                                        &25.6      \\ 
DDETR~\cite{zhu2020deformable}                                                             & \multicolumn{1}{c|}{0.0568}            & \multicolumn{1}{c|}{32168}            & 60.4                                                        & \multicolumn{1}{c|}{0.0260}            & \multicolumn{1}{c|}{10915}            & \multicolumn{1}{c|}{3.9}                                                           & \multicolumn{1}{c|}{38.0}                                                        & 21.0     & \multicolumn{1}{c|}{0.0155}            & \multicolumn{1}{c|}{7352}            & \multicolumn{1}{c|}{1.2}                                                           & \multicolumn{1}{c|}{26.5}                                                        & 9.6    & \multicolumn{1}{c|}{1.8}                                                           & \multicolumn{1}{c|}{22.5}                                                        & 7.0    \\ 
\begin{tabular}[c]{@{}c@{}}DDETR+Finetuning\end{tabular}       & \multicolumn{1}{c|}{-}            & \multicolumn{1}{c|}{-}            & -                                                        & \multicolumn{1}{c|}{0.0330}            & \multicolumn{1}{c|}{17260}            & \multicolumn{1}{c|}{54.1}                                                           & \multicolumn{1}{c|}{36.1}                                                        & 45.1     & \multicolumn{1}{c|}{0.0187}            & \multicolumn{1}{c|}{10432}            & \multicolumn{1}{c|}{40.1}                                                           & \multicolumn{1}{c|}{24.0}                                                        & 34.7     & \multicolumn{1}{c|}{33.1}                                                           & \multicolumn{1}{c|}{19.9}                                                        &29.8      \\ \hline
ORE-EBUI~\cite{joseph2021towards}$^{\dag}$                                                          & \multicolumn{1}{c|}{0.0621}            & \multicolumn{1}{c|}{10459}            & 56.0                                                        & \multicolumn{1}{c|}{0.0282}            & \multicolumn{1}{c|}{10445}            & \multicolumn{1}{c|}{52.7}                                                           & \multicolumn{1}{c|}{26.0}                                                        & 39.4     & \multicolumn{1}{c|}{0.0211}            & \multicolumn{1}{c|}{7990}            & \multicolumn{1}{c|}{38.2}                                                           & \multicolumn{1}{c|}{12.7}                                                        &29.7      & \multicolumn{1}{c|}{29.6}                                                           & \multicolumn{1}{c|}{12.4}                                                        &25.3      \\ 
OW-DETR~\cite{gupta2021ow}                                                           & \multicolumn{1}{c|}{0.0571}            & \multicolumn{1}{c|}{10240}            & 59.2                                                        & \multicolumn{1}{c|}{0.0278}            & \multicolumn{1}{c|}{8441}            & \multicolumn{1}{c|}{\textbf{53.6}}                                                           & \multicolumn{1}{c|}{33.5}                                                        &42.9      & \multicolumn{1}{c|}{0.0156}            & \multicolumn{1}{c|}{6803}            & \multicolumn{1}{c|}{38.3}                                                           & \multicolumn{1}{c|}{15.8}                                                        & 30.8     & \multicolumn{1}{c|}{31.4}                                                           & \multicolumn{1}{c|}{17.1}                                                        & 27.8     \\ 
Ours                                                              & \multicolumn{1}{c|}{\textbf{0.0549}}            & \multicolumn{1}{c|}{\textbf{5909}}            & \textbf{ 59.9}                                                       & \multicolumn{1}{c|}{\textbf{0.0210}}            & \multicolumn{1}{c|}{\textbf{4378}}            & \multicolumn{1}{c|}{51.8}                                                           & \multicolumn{1}{c|}{\textbf{36.4}}                                                        &\textbf{44.1}      & \multicolumn{1}{c|}{\textbf{0.0133}}            & \multicolumn{1}{c|}{\textbf{2895}}            & \multicolumn{1}{c|}{\textbf{38.9}}                                                           & \multicolumn{1}{c|}{\textbf{24.7}}                                                        & \textbf{34.2}     & \multicolumn{1}{c|}{\textbf{32.0}}                                                           & \multicolumn{1}{c|}{\textbf{19.7}}                                                        &\textbf{29.0}      \\ \hline
\end{tabular}}
\caption{\small Comparisons on MS-COCO. Results of methods marked $\dag$ are from OW-DETR~\cite{gupta2021ow}. Task 1: No fine-tuning since no previous known classes. Task 4: WI $\&$ A-OSE not computed since all 80 classes are known.} 
\label{results 2}
\end{table*}

\begin{table*}[!htbp]
\Large
\centering
\resizebox{\linewidth}{!}{
\begin{tabular}{c|cccccccccccccccccccc|c}
\hline
10+10 setting                                               & aero & cycle & bird & boat & bottle & bus & car & cat & chair & cow & table & dog & horse & bike & person & plant & sheep & sofa & train & tv & mAP \\ \hline
ILOD~\cite{shmelkov2017incremental}                                                        & 69.9     & 70.4      &69.4      & 54.3     & 48.0       & 68.7    & 78.9    & 68.4    & 45.5      & 58.1    & 59.7      &  72.7   &73.5       & 73.2     & 66.3       & 29.5      & 63.4      &  61.6    & 69.3      & 62.2   & 63.2    \\ 
Faster ILOD~\cite{peng2020faster}                                                & 72.8     & 75.7      & 71.2     &60.5      & 61.7       &70.4     & 83.3    & 76.6    & 53.1      & 72.3    & 36.7      & 70.9    & 66.8      & 67.6     &  66.1      &  24.7     & 63.1      &  48.1    & 57.1      &43.6    &62.1     \\ 
\begin{tabular}[c]{@{}c@{}}ORE - (CC+EBUI)\end{tabular}~\cite{joseph2021towards}   & 53.3     & 69.2      & 62.4     & 51.8     & 52.9       &73.6     & 83.7    & 71.7    & 42.8      &  66.8   & 46.8      & 59.9    & 65.5      &  66.1    & 68.6       & 29.8      &  55.1     & 51.6     & 65.3      & 51.5   & 59.4    \\ 
ORE-EBUI~\cite{joseph2021towards}                                                    & 63.5      &  70.9    & 58.9    &42.9  &34.1     &76.2     & 80.7    & 76.3    & 34.1      &66.1     & 56.1      &  70.4   &80.2       & 72.3     & 81.8       &  42.7     &71.6       & 68.1     & 77.0      &67.7    & 64.5   \\ 
OW-DETR~\cite{gupta2021ow}                                                     & 61.8     & 69.1      &  67.8    & 45.8     & 47.3       &78.3     & 78.4    &  78.6   &  36.2     & 71.5    & 57.5      &  75.3   & 76.2      & 77.4     & 79.5       & 40.1      & 66.8      & 66.3     & 75.6      & 64.1   &  65.7   \\\hline 
Ours                                                        &76.6      & 69.3      &75.2      &59.0      & 47.2       & 71.8    & 84.9    &81.0     & 42.7      & 70.5    & 69.7      & 80.7    &  83.4     & 69.0     &  83.1      &  38.9     & 65.5      & 66.8     & 77.9      & 73.7   &\textbf{69.3}     \\ \hline \hline
15+5 setting                                               & aero & cycle & bird & boat & bottle & bus & car & cat & chair & cow & table & dog & horse & bike & person & plant & sheep & sofa & train & tv & mAP \\ \hline
ILOD~\cite{shmelkov2017incremental}                                                        & 70.5     & 79.2      & 68.8     & 59.1     & 53.2       &75.4     & 79.4    & 78.8    &46.6       & 59.4    & 59.0      & 75.8    & 71.8      & 78.6     &  69.6      &33.7       & 61.5      & 63.1     &71.7       &62.2    & 65.8    \\ 
Faster ILOD~\cite{peng2020faster}                                                 &66.5      & 78.1      &71.8      &54.6      & 61.4       & 68.4    & 82.6    & 82.7    & 52.1      & 74.3    & 63.1      & 78.6    & 80.5      &  78.4    &80.4        & 36.7      & 61.7      & 59.3     & 67.9      & 59.1   & 67.9   \\ 
\begin{tabular}[c]{@{}c@{}}ORE - (CC+EBUI)\end{tabular}~\cite{joseph2021towards}  & 65.1     & 74.6      &57.9      & 39.5     & 36.7       &75.1     & 80.0    & 73.3    & 37.1      & 69.8    &  48.8     & 69.0    &  77.5     & 72.8     & 76.5       &34.4       & 62.6      & 56.5     & 80.3      & 65.7   & 62.6   \\ 
ORE-EBUI~\cite{joseph2021towards}                                                    & 75.4     & 81.0      & 67.1     &51.9      & 55.7       &77.2     & 85.6    & 81.7    &46.1       &  76.2   &  55.4     & 76.7    & 86.2      & 78.5     & 82.1       &  32.8     & 63.6      &  54.7    &  77.7     & 64.6   & 68.5    \\ 
OW-DETR~\cite{gupta2021ow}                                                     & 77.1     & 76.5      &  69.2    & 51.3     &  61.3      & 79.8    & 84.2    & 81.0    & 49.7      & 79.6    & 58.1      & 79.0    & 83.1      & 67.8     & 85.4       &  33.2     &  65.1     & 62.0     & 73.9      & 65.0   & 69.4    \\\hline 
Ours                                                        &81.4      & 81.1      & 76.8     &52.6      & 59.4       & 78.9    & 88.7    &88.5     & 54.7      & 70.6    & 49.9      &  84.3   & 87.7      & 82.6     & 83.1       &  28.7     &  57.9     & 55.8     & 79.7      &  62.3  &\textbf{70.2}     \\ \hline \hline
19+1 setting                                               & aero & cycle & bird & boat & bottle & bus & car & cat & chair & cow & table & dog & horse & bike & person & plant & sheep & sofa & train & tv & mAP \\ \hline
ILOD~\cite{shmelkov2017incremental}                                                         & 69.4     & 79.3      & 69.5     &57.4      &  45.4      & 78.4    &  79.1   &80.5     & 45.7      & 76.3    & 64.8      &  77.2   & 80.8      & 77.5     &  70.1      &42.3       & 67.5      & 64.4     &  76.7     &62.7    & 68.2    \\ 
Faster ILOD~\cite{peng2020faster}                                                 &  64.2    & 74.7      & 73.2     & 55.5     &  53.7      & 70.8    & 82.9    & 82.6    &51.6       & 79.7    &  58.7     & 78.8    &  81.8     & 75.3     & 77.4       &  43.1     & 73.8      & 61.7     &   69.8    & 61.1   &68.5    \\ 
\begin{tabular}[c]{@{}c@{}}ORE - (CC+EBUI)\end{tabular}~\cite{joseph2021towards}   &60.7      & 78.6      & 61.8     & 45.0     & 43.2       &   75.1  & 82.5    & 75.5    & 42.4      &  75.1   &  56.7     &  72.9   & 80.8      & 75.4     &  77.7      & 37.8      &  72.3     & 64.5     &  70.7     & 49.9   &  64.9   \\ 
ORE-EBUI~\cite{joseph2021towards}                                                    & 67.3     & 76.8      &60.0      & 48.4     & 58.8       &81.1 & 86.5    & 75.8    & 41.5      & 79.6    &  54.6     & 72.8    & 85.9      &  81.7    & 82.4       &  44.8     & 75.8      & 68.2     & 75.7      & 60.1   & 68.8    \\ 
OW-DETR~\cite{gupta2021ow}                                                    & 70.5     & 77.2      & 73.8     & 54.0     &55.6        &  79.0   &  80.8   & 80.6    & 43.2      & 80.4    & 53.5      & 77.5    & 89.5      & 82.0     &  74.7      & 43.3      &  71.9     & 66.6     & 79.4      & 62.0   & 70.2    \\ \hline
Ours                                                              & 80.1    & 84.7    & 78.7    & 59.3      & 62.8    & 81.7      &89.2     & 88.9      &55.5      & 82.2       & 68.6      &  86.4     & 87.3     & 83.1      & 83.9   & 47.3   &  76.0    &  71.8     &  88.0    & 64.4     &\textbf{76.0}   \\ \hline 
\end{tabular}}
\caption{\small Comparison with state-of-the-art for incremental object detection.} \vspace{-3mm}
\label{incremental setting}
\end{table*}

\vspace{-3mm}
\paragraph{Implementation details.}
We use the standard Deformable DETR~\cite{zhu2020deformable} object detector with a ResNet-50 backbone pre-trained on ImageNet~\cite{deng2009imagenet} in a self-supervised way~\cite{caron2021emerging}. 
The network architectures and hyperparameters of the transformer encoder and decoder remain the same as Deformable DETR.
The training is done on 8 RTX 3090 GPUs with a batch size of 2 per GPU.
In the model pre-training stage, we train our model using the AdamW~\cite{kingma2014adam,loshchilov2017decoupled} optimizer with an initial learning rate of $2 \times 10^{-4}$ and a weight decay of $1 \times 10^{-4}$. We train our model for 50 epochs and the learning rate is decayed at 40$^\textup{{th}}$ epoch by a factor of $0.1$. 
In the open world learning stage, the model is initialized from the pre-trained model. The parameters of the binary classification head, the class-specific projection layer and classification head are then fine-tuned while keeping the other parameters frozen. We fine-tune the model for 5 epochs with a learning rate of $2 \times 10^{-4}$, which is decayed at 3$^\textup{{th}}$ epoch by a factor of $0.1$. Following~\cite{gupta2021ow}, the top $50$ high scoring predictions per image are used for evaluation during inference. The hyperparameters $\lambda_\text{b\_cls}$ and $\lambda_\text{con}$ are set to 1 and 1, respectively. 
For the incremental object detection, we follow the same training details as the open world object detection.

\vspace{-1mm}
\subsection{Open World Object Detection}
Table~\ref{results 1} shows the comparisons between our proposed method with the recently proposed ORE~\cite{joseph2021towards} and OW-DETR~\cite{gupta2021ow} for open world object detection. Following OW-DETR~\cite{gupta2021ow} and for a fair comparison, we make comparisons with results of ORE implemented by OW-DETR~\cite{gupta2021ow}, \ie ORE-EBUI, where the energy-based unknown identifier (EBUI) that relies on a held-out validation set with supervision of unknown classes is not utilized. The comparisons are done on known classes mAP and unknown classes recall (U-Recall).
Our proposed method significantly outperforms the state-of-the-art method, OW-DETR, in terms of U-Recall on Task 1, 2 and 3, respectively. OW-DETR achieves U-Recall of 7.5, 6.2 and 5.7 on Task 1, 2 and 3, respectively. Our method improves the recall rate of the unknown classes, achieving U-Recall of 21.0, 15.7 and 17.4 on the Task 1, 2 and 3, respectively. Furthermore, our method outperforms OW-DETR in terms of mAP on all the four tasks for previously known and current known classes, except for the result of previously known classes on Task 2.
From these results, we can conclude that our method is effective for open world object detection with higher accuracy and recall rate.

\begin{figure*}[!htbp]
\centering
\includegraphics[width=\linewidth,height=4.5cm]{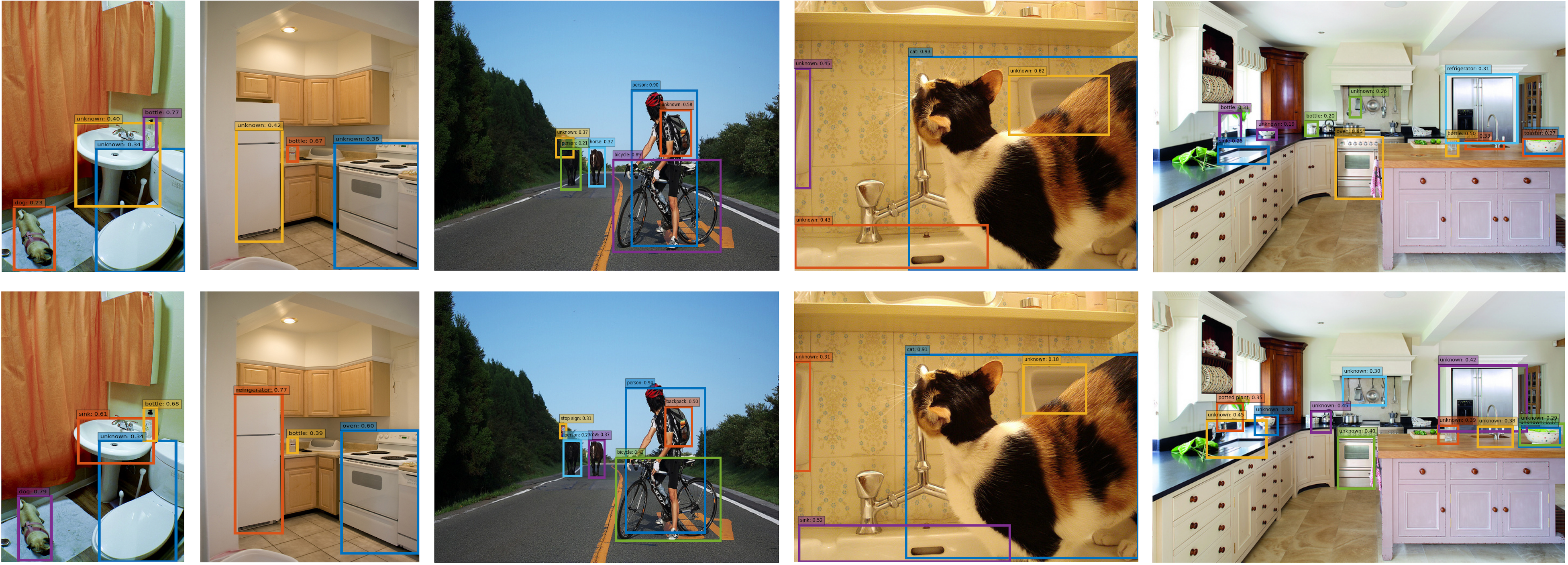}
\caption{Qualitative results of the proposed Open World DETR on MS-COCO $val$ set. 
} \vspace{-3mm}
\label{qualitative results}
\end{figure*}

\begin{table*}[!htbp]
\Large
\centering
\resizebox{\linewidth}{!}{
\begin{tabular}{p{1.5cm}<{\centering}|p{3.5cm}<{\centering}|p{1.5cm}<{\centering}p{3.5cm}<{\centering}|p{2.5cm}<{\centering}|p{1cm}<{\centering}p{2cm}<{\centering}|p{1cm}<{\centering}p{1cm}<{\centering}p{1cm}<{\centering}p{1cm}<{\centering}}

\hline
\multirow{3}{*}{Row ID} & \multirow{3}{*}{\begin{tabular}[c]{@{}c@{}}Two-stage \\ learning pipeline\end{tabular}} & \multicolumn{2}{c|}{Multi-view self-labeling strategy}                                       & \multirow{3}{*}{\begin{tabular}[c]{@{}c@{}}Consistency \\constraint\end{tabular}} & \multicolumn{2}{c|}{Task 1}                                                                                                                     & \multicolumn{4}{c}{Task 2}     \\ \cline{3-4} \cline{6-11} 
                        &                                              & \multicolumn{1}{c|}{\multirow{2}{*}{Binary classifier}} & \multirow{2}{*}{Selective search} &                              & \multicolumn{1}{c|}{U-Recall} &\multicolumn{1}{c|}{mAP($\uparrow$)}                                                  & \multicolumn{1}{c|}{U-Recall} & \multicolumn{3}{c}{mAP($\uparrow$)}                                                                                                                                             \\ 
                        &                                              & \multicolumn{1}{c|}{}                                   &                                   &                                & \multicolumn{1}{c|}{($\uparrow$)}       & \begin{tabular}[c]{@{}c@{}}Current \\ known\end{tabular}    & \multicolumn{1}{c|}{($\uparrow$)}       & \multicolumn{1}{c}{\begin{tabular}[c]{@{}c@{}}Previously\\ known\end{tabular}} & \multicolumn{1}{c}{\begin{tabular}[c]{@{}c@{}}Current \\ known\end{tabular}} & Both \\ \hline

1                       &    \checkmark                                          & \multicolumn{1}{c|}{\checkmark}                                   &                                   &                                 & \multicolumn{1}{c|}{18.1}         & 59.4                                                              & \multicolumn{1}{c|}{13.5}         & \multicolumn{1}{c|}{51.3}                                                           & \multicolumn{1}{c|}{35.7}                                                         & 43.5     \\ 
2                       &   \checkmark                                           & \multicolumn{1}{c|}{\checkmark}                                   & \checkmark                                  &                                   & \multicolumn{1}{c|}{\textbf{21.3}}         &  59.6                                                        
     & \multicolumn{1}{c|}{14.4}         & \multicolumn{1}{c|}{51.5}                                                           & \multicolumn{1}{c|}{35.7}                                                         & 43.6 
   \\ 
3                       &                                             & \multicolumn{1}{c|}{\checkmark}                                   & \checkmark                                   &  \checkmark                                  & \multicolumn{1}{c|}{19.9}         &   55.3                                                          & \multicolumn{1}{c|}{13.3}         & \multicolumn{1}{c|}{50.2}                                                           & \multicolumn{1}{c|}{34.3}                                                         & 42.3     \\ 
4                      & \checkmark                                             & \multicolumn{1}{c|}{\checkmark}                                   & \checkmark                                  &    \checkmark                             & \multicolumn{1}{c|}{21.0}         &   \textbf{59.9}                                                          & \multicolumn{1}{c|}{\textbf{15.7}}         & \multicolumn{1}{c|}{\textbf{51.8}}                                                           & \multicolumn{1}{c|}{\textbf{36.4}}                                                         & \textbf{44.1}     \\ \hline
\end{tabular}}
\caption{\small Ablation study of each component in the proposed Open World DETR.} 
\label{ablation}
\end{table*}

As shown in Table~\ref{results 2}, we also make comparisons in terms of wilderness impact (WI), absolute open set error (A-OSE) and known class mAP, which are previously reported in ORE~\cite{joseph2021towards}. We can see that our method has significantly lower WI and A-OSE scores on Task 1, 2 and 3 due to the better capability of our method on distinguishing unknown objects from objects of the known classes. Particularly on A-OSE, our method achieves scores close to half of the state-of-the-art OW-DETR on Task 1 (Our: 5909 vs. OW-DETR: 10240) and Task 2 (Ours: 4378 vs. OW-DETR: 8441), and even close to one-third of OW-DETR on Task 3 (Ours: 2895 vs. OW-DETR: 6803). These results show the superiority of our method in handling unknown objects.

Figure~\ref{qualitative results} shows several qualitative results of our Open World DETR on MS-COCO $val$ set. 
The top row shows the objects detected by our method when trained with annotations of Task 1 classes. 
We can see that unknown objects whose annotations are not given can be successfully classified as unknown.
The bottom row shows the predictions for the same images after incremental training with annotations of Task 2 classes.
We can see that unknown objects whose annotations are given on Task 2 can be correctly detected as one of the known classes. The unknown objects whose annotations are not given are still detected as unknown.

\vspace{-1mm}
\subsection{Incremental Object Detection}
Following~\cite{joseph2021towards,gupta2021ow}, we also do incremental learning on base classes and novel classes that are added incrementally.
We conduct experiments on PASCAL VOC 2007 that contains objects from 20 different classes.
The evaluations are performed on three standard settings, where we take the first 10, 15 and 19 classes of PASCAL VOC as the base classes, and the remaining 10, 5 and last class are used as novel classes, respectively. 
As shown in Table~\ref{incremental setting}, our method performs well on the incremental setting. This is because our model is capable of reducing the confusion of an unknown object being mistakenly classified as one of the known classes. Furthermore, the detection of unknown classes improves the capability of the model in adapting to novel classes.
We can see that our method significantly outperforms the state-of-the-art on all the three settings, which demonstrates the superiority of our method.

\vspace{-1mm}
\subsection{Ablation Study}
Table~\ref{ablation} shows the ablation studies to understand the effectiveness of each component in our method.  The ablated components include: 1) two-stage learning pipeline; 2) multi-view self-labeling strategy; 3) consistency constraint. 

Row 1 shows that by using the pseudo ground truths generated by the binary classifier in a swapped prediction mechanism, our method already achieves better performance than OW-DETR. This indicates training on these pseudo ground truths enables the capability of the model in detecting unknown objects.
We also use the proposals generated by the selective search as additional pseudo ground truths when these proposals have no overlaps with the pseudo ground truths generated by the binary classifier and no overlaps with the ground truths of the known classes. 
We can see from Row 2 that the using of the additional pseudo ground truths generated by selective search improves the performance of only using the binary classifier.
Furthermore, adding consistency constraint to this configuration in Row 4 gives the best performance, with the exception of a marginal drop in U-Recall of Task 1. It indicates that applying consistency constraint on object query features does help the learning of representations.

In Row 3, there is a severe performance drop in all terms compared with our final framework in Row 4. We think this largely shows that the two-stage learning pipeline is crucial for keeping the performance of known classes and learning to detect unknown classes.  
Open world object detection aims to detect the unknown objects as objects, which is contradictory to the training of the known classes that tends to detect unknown objects as background.
Concurrently, the pseudo ground truths of unknown classes contain many false positives. When the class-agnostic components are not kept frozen, the detection of the unknown objects enforces the feature space of the model to deviate severely from the pre-trained model.
These results show that each component in our method has a critical role for open world object detection.

\vspace{-2mm}
\section{Conclusion}
In this paper, we propose the Open World DETR for the more challenging and realistic scenario of
open world object detection.
Our Open World DETR is a two-stage training approach. 
In the first stage, we train Deformable DETR and an additional binary classifier to build unbiased representations. In the second stage, we introduce a multi-view self-labeling strategy to generate high-quality pseudo proposals for the unknown classes using the pre-trained class-agnostic binary classifier and the selective search algorithm.
We further enforce a consistency constraint on the object query features to enhance the quality of the representations.
Extensive experimental results on benchmark datasets show the effectiveness of our proposed method.
We hope this work can provide good insights and inspire further researches in the relatively under-explored but important problem of open world object detection.

\begin{figure*}[!htbp]
\centering
\includegraphics[width=0.95\textwidth,height=5.5cm]{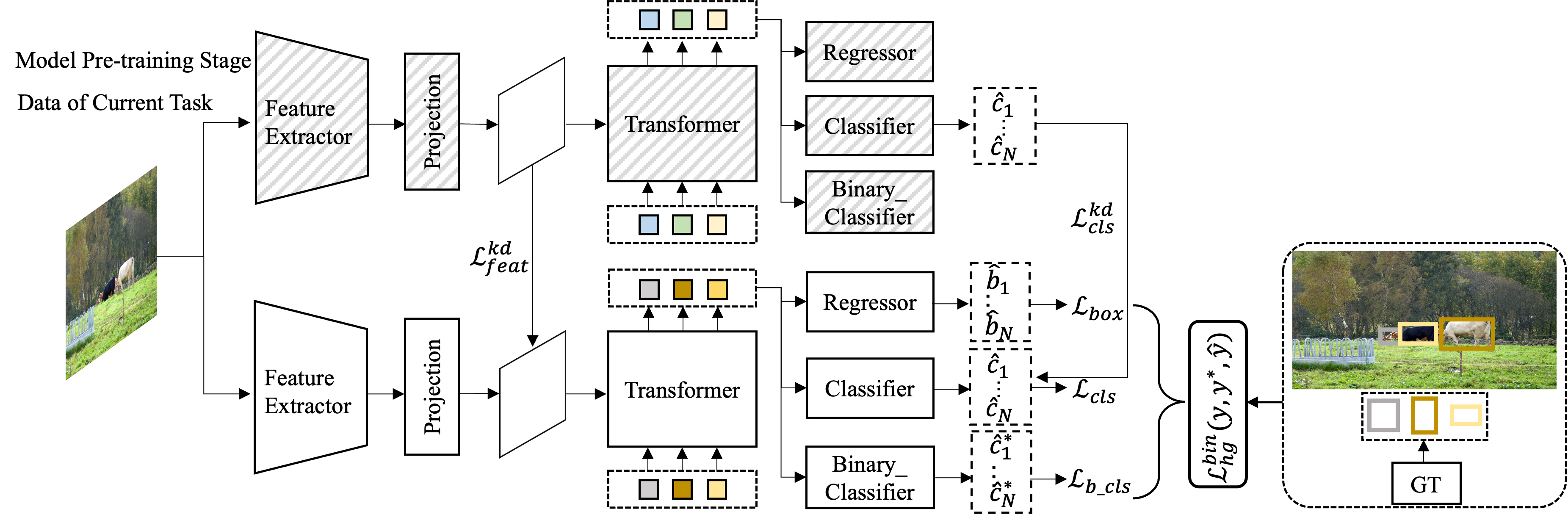}
\caption{Overview of our proposed model pre-training stage. Parameters of the modules shaded in gray slash line are frozen during training. Refer to the text for more details.}
\label{stage 1}
\end{figure*}


\newpage
\appendix
\section{Additional Details}

\subsection{Model Pre-training Stage}

\begin{figure*}[!t]
\centering
\includegraphics[width=\linewidth,height=12.0cm]{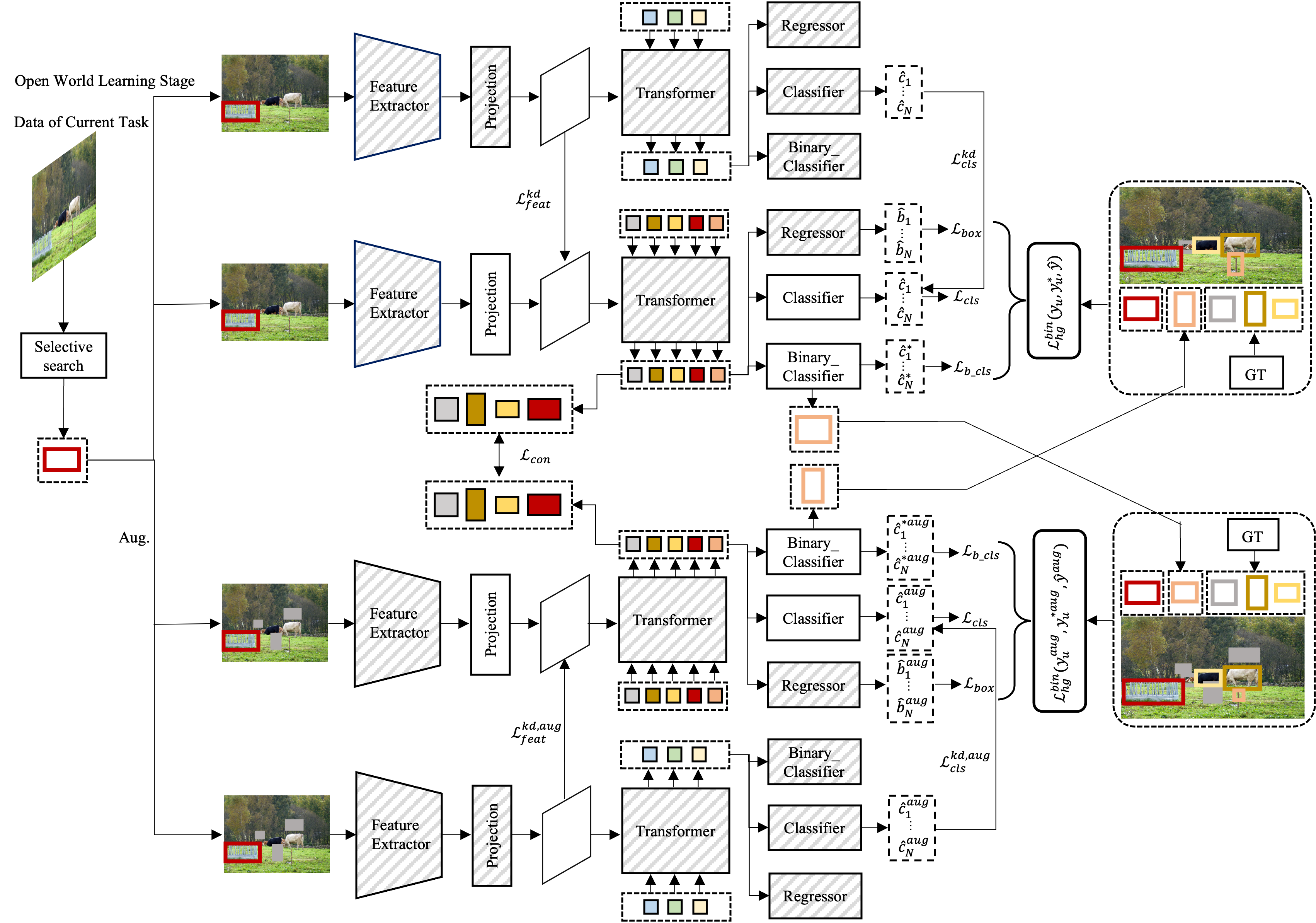}
\caption{Overview of our proposed open world learning stage. Parameters of the modules shaded in gray slash line are frozen during training. Refer to the text for more details.
} 
\label{stage 2}
\end{figure*} 

Task 1 only requires the detection of unknown objects as objects without any previous known class. 
In addition to the detection of unknown objects as objects, the model is required to mitigate catastrophic forgetting of the previous known classes when trained only on the dataset with annotations of the current known classes on Task 2, 3 and 4.  
On Task 2, 3 and 4, the current model is initialized from the previous model. 
However, the constant updating of the parameters of the current model can aggravate catastrophic forgetting of the previous known classes. To circumvent this problem, we propose to use knowledge distillation to alleviate catastrophic forgetting.

Figure~\ref{stage 1} shows our first stage model pre-training on Task 2, 3 and 4. 
The previous model is employed as an extra supervision signal to prevent significant deviation among the output features of the previous and current models.
Since the projection layer is class-specific and the backbone ResNet is class-agnostic, our work applies knowledge distillation on the output features of the class-specific projection layer instead of the output features of the backbone as previous works. Consequently, we can effectively alleviate the forgetting of the previous known classes after the class-specific projection layer. 
Furthermore, a direct knowledge distillation on the full features causes conflicts and thus impedes the learning of the current known classes. 
We thus use the ground truth bounding boxes of the current known classes to form a binary mask $\mathit{mask}$ to prevent negative influence on the current known class learning from the features of the previous model.  Specifically, we set the value of the pixel on the feature map within the ground truth bounding boxes of the current known classes as 1, and the value of  the pixel outside the ground truth bounding boxes as 0.
The distillation loss with the mask on the features is written as:
\begin{equation}
\small
    \mathcal{L}_\text{feat}^\text{kd} =  \frac{1}{2N} \sum_{i=1}^ w \sum_{j=1}^ h \sum_{k =1} ^c (1 - \mathit{mask}_{ij}) \Big\|  \mathit{f}^\text{cur}_{ijk} - \mathit{f}^\text{pre}_{ijk} \Big\| ^2,
\end{equation}
where  $N = \sum_{i=1}^ w \sum_{j=1}^ h (1 - \mathit{mask}_{ij})$. 
$\mathit{f^\text{pre}}$ and $\mathit{f^\text{cur}}$  denote the feature of the previous and current models, respectively. $\mathit{w}$, $\mathit{h}$ and $\mathit{c}$ are the width, height and channels of the feature map. 

In addition, knowledge distillation at the classification head is also adopted.
We first select the prediction outputs from the $N$ prediction outputs of the previous model as pseudo ground truths of the previous known classes.
Specifically, for an input image, we consider a prediction output of the previous model as a pseudo ground truth of the previous known classes when its class probability is greater than threshold 0.5 and its bounding box has no overlap with ground truth bounding boxes of the current known classes.  
We then adopt a pair-wise matching cost to find the bipartite matching between the pseudo ground truths and the predictions of the current model. Subsequently, the classification outputs of the previous and current models are compared in the distillation loss function given by:
\begin{equation}
\begin{array}{rll}
\begin{aligned}
\displaystyle 
\mathcal{L}_\text{cls}^\text{kd} &= 
 \mathcal{L}_\text{kl\_div} ( \log (\mathit{p^\text{cur}}), \mathit{p^\text{pre}}),  \\
\end{aligned}
\end{array}
\end{equation}
where we follow \cite{hinton2015distilling} in the definition of the KL-divergence loss $\mathcal{L}_\text{kl\_div}$ between the class probabilities of the current and previous models. $\mathit{p}$ denotes the class probability.

The overall loss $\mathcal{L}_\text{overall}^\text{pt}$ of our proposed model pre-training stage on Task 2, 3 and 4 is given by:
\begin{equation}
\begin{array}{rll}
\begin{aligned}
\displaystyle 
\mathcal{L}_\text{overall}^\text{pt} &= \mathcal{L}^\text{bin}_\text{hg} (y, y^*, \hat{y}) + \lambda_\text{feat}\mathcal{L}_\text{feat}^\text{kd} + \lambda_\text{cls}\mathcal{L}_\text{cls}^\text{kd},
\end{aligned}
\end{array}
\end{equation}
where $\lambda_\text{feat}$ and $\lambda_\text{cls}$ are hyperparameters to balance the loss terms.  $\mathcal{L}^\text{bin}_\text{hg} (y, y^*, \hat{y})$ (\cf Eq.~4 of the main paper) is the Hungarian loss for all pairs matched of the traditional classification head, the regression head and the binary classification head.

\begin{table*}[!htbp]
\Large
\centering
\resizebox{\linewidth}{!}{
\begin{tabular}{c|cccccccccccccccccccc|c}
\hline

19+1 setting                                               & aero & cycle & bird & boat & bottle & bus & car & cat & chair & cow & table & dog & horse & bike & person & plant & sheep & sofa & train & tv & mAP \\ \hline

Ours - owod                                                    & 76.3     & 81.1      & 76.5     & 54.9     &59.2        &  80.7   &  88.0   & 85.9    & 51.5      & 79.7    & 64.8      & 83.0   & 84.9     & 79.0    &  79.7      & 42.1      &  72.1     & 70.4     & 86.9      & 62.0   & 72.9    \\ \hline
Ours                                                              & 80.1    & 84.7    & 78.7    & 59.3      & 62.8    & 81.7      &89.2     & 88.9      &55.5      & 82.2       & 68.6      &  86.4     & 87.3     & 83.1      & 83.9   & 47.3   &  76.0    &  71.8     &  88.0    & 64.4     &\textbf{76.0}   \\ \hline
\end{tabular}}
\caption{\small Ablation study for Incremental Object Detection.} 
\label{incremental setting}
\end{table*}

\begin{table}[!t]
\Large
\centering
\resizebox{\linewidth}{!}{
\begin{tabular}{c|c|c|c|cc|c}
\hline
\multirow{3}{*}{Row ID}& \multirow{3}{*}{\begin{tabular}[c]{@{}c@{}} Knowledge \\ distillation \end{tabular}}& \multirow{3}{*}{\begin{tabular}[c]{@{}c@{}} Exemplar \\ replay \end{tabular}}                                                                                                             & \multicolumn{4}{c}{Task 2}                                                                                                                                                                                                                                   \\ \cline{4-7} 
            &   &             & \multicolumn{1}{c|}{U-Recall} & \multicolumn{3}{c}{mAP($\uparrow$)}                                                                                                                                             \\ 
                        &                                        &    & \multicolumn{1}{c|}{($\uparrow$)}       & \multicolumn{1}{c}{\begin{tabular}[c]{@{}c@{}}Previously\\ known\end{tabular}} & \multicolumn{1}{c}{\begin{tabular}[c]{@{}c@{}}Current \\ known\end{tabular}} & Both \\ \hline

1                  & \checkmark  &           & \multicolumn{1}{c|}{13.4}         & \multicolumn{1}{c|}{6.3}                                                           & \multicolumn{1}{c|}{\textbf{37.9}}                                                         & 22.1     \\ 
2                 &   & \checkmark            & \multicolumn{1}{c|}{\textbf{18.5}}         & \multicolumn{1}{c|}{33.7}                                                           & \multicolumn{1}{c|}{36.7}                                                         & 35.2     \\ 
3                   & \checkmark  & \checkmark       & \multicolumn{1}{c|}{15.7}         & \multicolumn{1}{c|}{\textbf{51.8}}                                                           & \multicolumn{1}{c|}{36.4}                                                        & \textbf{44.1}     \\ \hline
\end{tabular}}
\caption{\small Ablation study for the effectiveness of knowledge distillation and exemplar replay of Open World Object Detection.} 
\label{ablation_kd_ft}
\end{table}

\subsection{Open World Learning Stage}

Figure~\ref{stage 2} shows our second stage open world learning on Task 2, 3 and 4. To alleviate the problem of catastrophic forgetting of the previous known classes, we also apply knowledge distillation to the projection layer outputs and the classification outputs of images $\mathcal{I}$ and $\mathcal{I}^\text{aug}$, respectively. 
We use the ground truth bounding boxes of the current known classes of images $\mathcal{I}$ and $\mathcal{I}^\text{aug}$ to form binary masks $\mathit{mask}$ and  $\mathit{mask}^\text{aug}$ to prevent negative influence on the current known class learning from features of the previous model. The distillation losses  with the masks on the features of images $\mathcal{I}$ and $\mathcal{I}^\text{aug}$ are written respectively as:
\begin{equation}
\begin{array}{rll}
\begin{aligned}
\small
   & \mathcal{L}_\text{feat}^\text{kd} =  \frac{1}{2N} \sum_{i=1}^ w \sum_{j=1}^ h \sum_{k =1} ^c (1 - \mathit{mask}_{ij}) \Big\|  \mathit{f}^\text{cur}_{ijk} - \mathit{f}^\text{pre}_{ijk} \Big\| ^2, \\
    &\mathcal{L}_\text{feat}^\text{kd,aug} =  \frac{1}{2M} \sum_{i=1}^ w \sum_{j=1}^ h \sum_{k =1} ^c (1 - \mathit{mask}^\text{aug}_{ij}) \Big\|  \mathit{f}^\text{cur,aug}_{ijk} - \mathit{f}^\text{pre,aug}_{ijk} \Big\| ^2,
\end{aligned}
\end{array}
\end{equation}
where  $N = \sum_{i=1}^w \sum_{j=1}^h (1 - \mathit{mask}_{ij})$ and $M = \sum_{i=1}^w \sum_{j=1}^h (1 - \mathit{mask}^\text{aug}_{ij})$. 
$\mathit{f^\text{pre}}$ and $\mathit{f^\text{cur}}$  denote the feature of the previous and current models of image $\mathcal{I}$, respectively. $\mathit{f^\text{pre,aug}}$ and $\mathit{f^\text{cur,aug}}$  denote the feature of the previous and current models of image $\mathcal{I}^\text{aug}$, respectively. $\mathit{w}$, $\mathit{h}$ and $\mathit{c}$ are the width, height and channels of the feature map.

For an input image, we consider a prediction output of the previous model as a pseudo ground truth of the previous known classes when its class probability is greater than threshold 0.5 and its bounding box has no overlap with ground truth bounding boxes of the current known classes. Based on the pseudo ground truths, we adopt the pair-wise matching to search for the bipartite matching between the pseudo ground truths and the prediction outputs of the current model. 
Subsequently, the classification outputs of images $\mathcal{I}$ and $\mathcal{I}^\text{aug}$ from the previous and current models are compared in the following distillation losses respectively given by:
\begin{equation}
\begin{array}{rll}
\begin{aligned}
\displaystyle 
\mathcal{L}_\text{cls}^\text{kd} &= \mathcal{L}_\text{kl\_div} ( \log (\mathit{p^\text{cur}}), \mathit{p^\text{pre}}), \\
\mathcal{L}_\text{cls}^\text{kd,aug} &= 
 \mathcal{L}_\text{kl\_div} ( \log (\mathit{p^\text{cur,aug}}), \mathit{p^\text{pre,aug}}), 
\end{aligned}
\end{array}
\end{equation}
where $\mathit{p}$ denotes the class probability.

The overall loss $\mathcal{L}_\text{overall}^\text{owl}$ of our proposed open world learning stage on Task 2, 3 and 4 is given by:
\begin{equation}
\begin{array}{rll}
\begin{aligned}
\displaystyle 
\mathcal{L}_\text{overall}^\text{owl}& = \mathcal{L}_\text{total} 
 + \lambda_\text{feat}\mathcal{L}_\text{feat}^\text{kd} + \lambda_\text{cls}\mathcal{L}_\text{cls}^\text{kd}
 \\&+ \lambda_\text{feat}^\text{aug}\mathcal{L}_\text{feat}^\text{kd,aug} + \lambda_\text{cls}^\text{aug}\mathcal{L}_\text{cls}^\text{kd,aug},
\end{aligned}
\end{array}
\end{equation}
where $\lambda_\text{feat}$, $\lambda_\text{cls}$, $\lambda_\text{feat}^\text{aug}$ and $\lambda_\text{cls}^\text{aug}$ are hyperparameters to balance the loss terms. $\mathcal{L}_\text{total}$ (\cf Eq.~6 of the main paper) is the loss consists of: 1) Two Hungarian losses for all pairs matched between ground truths and predictions of the traditional classification head, the regression head and the binary classification head of images $\mathcal{I}$ and $\mathcal{I}^\text{aug}$; 
2) A consistency loss on the object query features of images $\mathcal{I}$ and $\mathcal{I}^\text{aug}$.

\begin{table}[!t]
\huge
\centering
\resizebox{\linewidth}{!}{
\begin{tabular}{c|ccccc|cc}
\hline
\multirow{2}{*}{Class} & \multicolumn{5}{c|}{Frozen layers}                                                                                                                     & \multicolumn{2}{c}{mAP}      \\ \cline{2-8} 
                       & \multicolumn{1}{c|}{Backbone} & \multicolumn{1}{c|}{Projection layer} & \multicolumn{1}{c|}{Transformer} & \multicolumn{1}{c|}{Regressor} & Classifier & \multicolumn{1}{c|}{Base} & Novel \\ \hline
1-41                   & \multicolumn{1}{c|}{-}        & \multicolumn{1}{c|}{-}                & \multicolumn{1}{c|}{-}           & \multicolumn{1}{c|}{-}         & -          & \multicolumn{1}{c|}{45.6} & 31.3  \\ 
1-40                   & \multicolumn{1}{c|}{-}        & \multicolumn{1}{c|}{-}                & \multicolumn{1}{c|}{-}           & \multicolumn{1}{c|}{-}         & -          & \multicolumn{1}{c|}{44.8} & -     \\ \hline 
\tabincell{c}{ 41 \\ (scratch) }                     & \multicolumn{1}{c|}{}        & \multicolumn{1}{c|}{}                & \multicolumn{1}{c|}{}           & \multicolumn{1}{c|}{}         &          & \multicolumn{1}{c|}{-}    & 16.7  \\ 
\tabincell{c}{41 \\ (fine-tune) }                       & \multicolumn{1}{c|}{ \checkmark}        & \multicolumn{1}{c|}{}                & \multicolumn{1}{c|}{}           & \multicolumn{1}{c|}{ \checkmark}         &           & \multicolumn{1}{c|}{-}    & 27.6  \\ 
\tabincell{c}{41 \\ (fine-tune) }                      & \multicolumn{1}{c|}{ \checkmark}        & \multicolumn{1}{c|}{}                & \multicolumn{1}{c|}{ \checkmark}           & \multicolumn{1}{c|}{ \checkmark}         &           & \multicolumn{1}{c|}{-}    & 30.7  \\ 
\tabincell{c}{41 \\ (fine-tune) }                       & \multicolumn{1}{c|}{ \checkmark}        & \multicolumn{1}{c|}{ \checkmark}                & \multicolumn{1}{c|}{ \checkmark}           & \multicolumn{1}{c|}{ \checkmark}         &           & \multicolumn{1}{c|}{-}    & 11.7  \\ \hline \hline
1-80                   & \multicolumn{1}{c|}{-}        & \multicolumn{1}{c|}{-}                & \multicolumn{1}{c|}{-}           & \multicolumn{1}{c|}{-}         & -          & \multicolumn{1}{c|}{46.8} & 36.3  \\ \hline

 \tabincell{c}{ 41-80 \\(scratch)}                     & \multicolumn{1}{c|}{}        & \multicolumn{1}{c|}{}                & \multicolumn{1}{c|}{}           & \multicolumn{1}{c|}{}         &         & \multicolumn{1}{c|}{-}    & 35.0  \\ 
\tabincell{c}{41-80 \\ (fine-tune)}                   & \multicolumn{1}{c|}{ \checkmark}        & \multicolumn{1}{c|}{}                & \multicolumn{1}{c|}{}           & \multicolumn{1}{c|}{ \checkmark}         &          & \multicolumn{1}{c|}{-}    & 33.2      \\ 
\tabincell{c}{41-80 \\ (fine-tune)}                   & \multicolumn{1}{c|}{ \checkmark}        & \multicolumn{1}{c|}{}                & \multicolumn{1}{c|}{ \checkmark}           & \multicolumn{1}{c|}{ \checkmark}         &           & \multicolumn{1}{c|}{-}    & 33.0  \\ 
\tabincell{c}{41-80 \\ (fine-tune)}                   & \multicolumn{1}{c|}{ \checkmark}        & \multicolumn{1}{c|}{ \checkmark}                & \multicolumn{1}{c|}{ \checkmark}           & \multicolumn{1}{c|}{ \checkmark}         &           & \multicolumn{1}{c|}{-}    & 17.8  \\ \hline
\end{tabular}}
\caption{\small Ablation study for the class-agnostic/-specific property of each module of Deformable DETR on COCO $val$ set. The layer with $\checkmark$ means this layer is frozen, otherwise this layer is optimized.} 
\label{frozen layers}
\end{table}

\section{Additional Experiments}

Table~\ref{incremental setting} shows the ablation studies to understand the effectiveness of our method in incremental object detection. The incremental object detection aims to achieve competitive results on novel classes and does not degrade the base class performance. Row 1 is the setting that only knowledge distillation and exemplar replay in our method are included to overcome the forgetting of base classes without using our proposed two-stage training approach. 
We can see that our method that comprises the proposed two-stage training approach significantly outperforms this setting (76.0\% mAP \vs 72.9\% mAP). 
This largely attribute to that the two-stage training approach does help the model well adapt to novel classes while maintain the knowledge of base classes.

We further validate the necessity of knowledge distillation and exemplar replay in open world object detection. As shown in Table~\ref{ablation_kd_ft},  we can see that using the knowledge distillation and fine-tuning with exemplar replay achieves the best result for both classes (44.1\% mAP) on Task 2. This result significantly outperforms the 22.1\% mAP when fine-tuning with exemplar replay is removed, and 35.2\% mAP when knowledge distillation is removed.

As shown in Table~\ref{frozen layers}, empirical experimental studies conducted to decide which module to freeze.  
We take the first 40 classes from MS COCO as the base classes, and the next 41$^\text{th}$ class is used as the novel class. Furthermore, we also take the first 40 classes from MS COCO as the base classes, and the next 40 classes as the additional group of novel classes.  When the novel class is the 41th class, the training data is scarce.  When the novel classes are the 41-80 classes, the training data is abundant. 
We pre-train the model on data of base classes. We initialize the novel model with the pre-trained base model when it is fine-tuned on data of novel classes.
The novel model achieves the best result of 30.7\% mAP when the projection layer and the classifier are not frozen under data scarcity. 
In the presence of abundance data, the novel model achieves 33.0\% mAP when the projection layer and classifier are not frozen. The above experiments can demonstrate that the projection layer and the classifier are class-specific and other modules are class-agnostic.

\section{Limitations and potential negative societal impacts}
 We observe that there are a lot of false positive proposals generated by our method for the unknown classes. We plan to use contrastive learning to mitigate this problem in our future work. We do not foresee any potential negative societal impacts by our work.

\clearpage

\end{document}